\documentclass[letterpaper]{article}
\usepackage{aaai}
\usepackage{times}
\usepackage{helvet}
\usepackage{courier}
\usepackage{graphicx}
\newcommand{\headingbf}[1]{\par\vspace{0.5ex}\noindent\textbf{#1}\par\vspace{0.25ex}}

\title{Design a Reliable LLM-Integrated Interface\\for Mortality Forecasting}
\author{Thi Kim Ngan Nguyen\\
Curtin University
}

\begin{document}
\nocopyright
\maketitle

\begin{abstract}
\begin{quote}
Mortality forecasting plays an important role in actuarial and policy decision-making, but its implementation remains technically complex and inaccessible to non-expert users. This project proposes a reliable large language model (LLM)-integrated interface that improves usability while maintaining statistical power. The LLM is designed as a constrained orchestration layer that translates natural-language inputs into structured configurations for a deterministic forecasting pipeline. A three-phase methodology is employed to ensure the accuracy, usability and transparency. First, a baseline pipeline is implemented using CoMoMo package, reproducing established mortality forecasting results. Second, the pipeline is extended to generate multi-step forecasts using rolling-origin evaluation and mean squared error (MSE). Third, a prototype interface using local LLM to handle users' forecasting requests in plain language. The system demonstrates that LLMs can enhance accessibility without compromising reproducibility, transparency, or actuarial validity in high-stakes analytical workflows.
\end{quote}
\end{abstract}

\section{Introduction}

\subsection{Background}
Over recent decades, mortality rate has changed significantly across populations. More specifically, life expectancy has been more than doubled from around 32 years in the early 1900s to approximately 71 years by 2021 (Dattani et al., 2023). In some developed countries, life expectancy of elderly people is even higher. Thus, understanding these trends has become more important for governments, pension systems or insurance companies to formulate policies. Consequently, mortality forecasting plays a key role in financial decisionmaking because small errors can lead to large long-term impacts on pension liabilities, life insurance pricing, and public-resource allocation.

As a result, there are many stochastic mortality models being introduced for past decades. For instance, there are six popular models: Lee-Carter (LC), Renshaw-Haberman (RH), Age-Period-Cohort (APC), CairnsBlake-Dowd (CBD), M7, and PLAT (Lee \& Carter, 1992; Renshaw \& Haberman, 2006; Cairns et al., 2006; Plat, 2009). Each model captures mortality dynamics differently, and each has strengths and weaknesses. Some models perform better for short-term forecasts, while others are more stable for longer horizons (Kessy et al., 2022). Therefore, selecting one single ``best'' model is not an ideal solution (Kessy et al., 2022; Shang \& Haberman, 2018).

To address this model-selection uncertainty, researchers have proposed model averaging approaches. For instance, Bayesian model averaging (BMA) which perform well in predicting trends, especially linearity, however, this technique is sensitive to single model's distribution and underperform in long forecast horizons (Shang and Haberman, 2018). Besides that, Kessy et al., 2022 employed stacked regression ensembles to tackle the limitations, this is a strong method because they learn data-driven horizon-specific weights to combine forecasts from multiple models. This reduces reliance on a single model and can improve out-of-sample accuracy.

\headingbf{Problem statement}
Though modern mortality models and ensemble methods are statistically powerful, their implementation is often complex when it comes to technical implementation. Building and comparing forecasting pipelines usually requires manual coding, learning specific software package conventions, and repeating complex configurations. Those setups are typically performed by experts in R or similar environments. That complexity excludes many non-technical users, and even for analysts who know how to program, it makes each new experiment a slow and error-prone undertaking. While the CoMoMo package was developed to implement stacked combination methods and mortality-prediction workflows in a more approachable way, it remains important to provide easier ways for users to configure and run forecasting studies without specialised R programming.

The research question addressed in this report is: How can we make mortality forecasting with CoMoMo package user-friendly through natural-language interaction, while preserving correctness, reproducibility and transparency?

\subsection{Motivation}
Large language models (LLMs) have demonstrated strong capability in translating natural language into structured tasks. This is a promising solution to bridge domain complexity and user accessibility. The motivation of this project is to apply an LLM as a natural language interaction that helps users create valid CoMoMo forecasting requests from plain English, without using the LLM as the forecasting model itself.

\subsection{Report structure}
We will present this report in 10 main parts. Research Objectives shows the research objectives and hypotheses. Significance discusses the significance of the study in relation to societal and professional actuarial practice. Literature review provides a literature review on mortality modelling, model combination, and LLM-enabled analytics. Methodology presents the methodology, including data sources, modelling approaches, system guardrails, and implementation design. Results and Discussion reports and analyses the empirical results from rolling-origin evaluation and multi-step forecasting outputs. Limitations discusses the limitations of the research. Future Work outlines directions for future work. Conclusion concludes the study by summarising the key findings. Finally, References lists the references, followed by appendices containing supplementary technical materials.

\section{Research Objectives}

\subsection{General objective}
The project generally aims to develop a reliable LLM-integrated mortality forecasting pipeline that can translate natural language requests into reproducible analytical outputs.

\subsection{Specific objective}

\headingbf{Correctness and reproducibility}
In this project, we replicate and verify a baseline by the deterministic CoMoMo pipeline under fixed effective configurations to establish accuracy benchmarks using rolling-origin mean squared error and model rankings. Additionally, we also quantify whether ensemble methods, including stacking, Bayesian model averaging, and model confidence set approaches, improve out-of-sample performance relative to individual models when evaluated under identical windows and implementations.

\headingbf{Usability}
An end-to-end prototype will be provided that can support natural-language or structured JSON input, connect to an API-driven back-end in R, and produce clearly defined artefacts via a Streamlit interface suitable for demonstration and employment by non-technical users.

\headingbf{Transparency}
To ensure the transparency of the model, model fitting, validation and testing outputs are presented such as mean squared error and evaluation summaries alongside forecasting results to provide clear performance baselines that help users to interpret, verify and assess the reliability of the forecasts.

\section{Significance}

\subsection{Societal Relevance}
It is unavoidable that understanding mortality trends is becoming more important in retirement security, healthcare planning, and social policy formulation. As a result, making models more accessible to stakeholders will support evidence-based decision-making faster in public organisations and private risk management contexts, without compromising analytical rigour.

\subsection{Industry Value}
For actuarial and risk analysts, the contribution of this project is practical. The proposed system supports auditable decision-making processes by reducing manual orchestration effort, accelerating model-comparison workflows, and generating traceable analytical artefacts. More importantly, the architecture emphasises deterministic statistical computation and transparency of professional-built models, rather than relying on AI generative outputs.

\subsection{Academic and Practical Value}
From an academic perspective, this work provides a foundation for future large language model applications in actuarial science. In practical terms, the project provides a reproducible design pattern for integrating LLMs into high-stakes analytics. The work directly reflects the proposal's core priorities: replicable evaluation, forecasting conducted within the same validated code path, and a demonstration-oriented front end that ensures usability improvements do not come at the expense of auditability or statistical integrity.

\section{Literature review}

\subsection{Stochastic Mortality Modelling Foundations}
Mortality forecasting has been deeply explored in actuarial science and demography, with stochastic modelling frameworks. For instance, one of the popular models is the Lee-Carter (LC) model, which uses the logarithm of mortality rates as a function of age-specific parameters and a dynamic mortality trend. As it is simple and interpretable, it is widely adopted in the field. However, this model does not work well when it comes to capture cohort effects and non-linear mortality trends. Thus, more flexible models have been developed within the Generalised Age-Period-Cohort (GAPC) framework. An extended version of Lee-Carter model is the Renshaw Haberman (RH) model, which incorporates cohort effects to better represent mortality patterns that commonly found in people in different age groups. Furthermore, the Age--Period--Cohort (APC) generalises this framework by explicitly modelling age, period and cohort components. More advanced models such as M7, add more terms to capture more complex relationships, including non-linear and age-dependent effects in mortality rates. A combination of Cairns-Blake-Dowd (CBD) with cohort effects and other factors is the PLAT model, which provide more flexibility and improved performance, long-term forecasting. These models have their own strengths and weaknesses which make it more complex to select one model only, as it may lead to suboptimal or unstable forecasts as it depends on context (Lee \& Carter, 1992; Renshaw \& Haberman, 2006; Cairns et al., 2006; Plat, 2009).

\subsection{Forecast Evaluation in Mortality Modelling}
Forecast evaluation is vital in mortality modelling because a model that fits historical data well may not be effective in producing accurate forecasts for future periods. In terms of model selection, some approaches which are commonly used are in-sample and out-of-sample evaluation, alongside techniques should also be considered are resampling and block cross-validation. In-sample fit refers to how well a model explains the data used to estimate its parameters. Although this is useful for understanding model adequacy, it may overestimate performance because the model is being assessed on the same distribution from which it learned its structure. In contrast, out-of-sample forecast accuracy is estimated based on unseen dataset, which makes it more relevant for forecasting where the main goal is to predict future mortality rates rather than describe past trends. For this reason, out-of-sample evaluation is considered more informative than in-sample fit when comparing mortality forecasting models (Hyndman \& Athanasopoulos, 2021).

In time-series forecasting, there is an approach called rolling-origin, the model is fitted on an initial training window, then used to produce forecasts for future periods. Then, the forecast origin will move step up and the process will keep going until it finishes. This will produce different forecast errors across periods and does not rely on a single train-test split. This evaluation method is suitable for mortality data as it respects the structure of series and more stable when assessing predictive performance across time. This illustrates how forecast actually works in real-time, when we make decisions based on past events.

\subsection{Selection Risk and Ensemble Approaches}
To tackle the concern of model selection risk, there are numerous studies on model combination techniques. The main idea is to combine multiple models to produce more robust and accurate results than a single model. Besides some traditional Simple Model Averaging, Bayesian Model Averaging, and Model Confidence Set, there is the method called Stacked Regression Ensemble which Kessy et al. (2022) applied to mortality forecasting. This ensemble combines the mortality rate forecasts from base learners using linear regression for single mortality model forecasts. There are 2 steps main steps, which is single models will generate their own cross-validated prediction, following that, a meta-learner is trained to learn the weights for combining multiple base learners which is helpful in enhancing predictive performance (Wolpert, 1992; Yao et al., 2018; Kessy et al., 2022).

\subsection{LLMs and their application in actuarial and healthcare analytics}
Actuarial science, or more specifically, mortality prediction has achieved a high level of sophistication in statistical modelling. While this is a strength, it also creates a barrier for non-expert users due to the complexity of implementation and interpretation. In recent years, Large Language Models (LLMs) have been built as powerful tools for natural language processing and human-computer interaction. LLMs have proven their strong capabilities in tasks like code generation, information extraction, and translation of unstructured text into structured outputs (e.g. JSON) (Brown et al., 2020; OpenAI, 2023).

Applications of LLMs in actuarial science and healthcare is mainly focusing on summarising reports, answering domain-specific questions, and extracting relevant information from large datasets. Though this is helpful in productivity improvements, LLMs are working as a single-layer system when they scan texts and analyse but not be integrated into established modelling pipelines. This limitation is considered to be important in high-stakes domains such as mortality forecasting, in which accuracy and reliability play a vital role.

Consequently, LLMs might lead to inconsistent outputs and occasional hallucinations as they have probabilistic nature. While their capability to engage with complex tasks is improving, their application in structured analytical domains is still controversial. As a result, recent research has increasingly positioned LLMs as orchestration or interface layers rather than primary analytical engines. In this case, LLMs translate user intent into structured commands that are executed by deterministic systems. This approach on one hand helps organisations to leverage the usability and accessibility benefits of LLMs while preserving the accuracy, reliability, and reproducibility of established modelling frameworks.

\subsection{Reliability concerns in high-stakes AI usage}
There are worrisome concerns regarding reliability and transparency being raised in employing artificial intelligence in high-stakes domains, such as actuarial science and healthcare. Since the large language models have the probabilistic nature, which means with the identical inputs, they may produce different outputs when executed. This variability causes inconsistency and risk in critical contexts.

Another challenge of LMs is the hallucination, which should be dealt with cautiously as they can generate plausible but incorrect information. In mortality forecasting, these types of errors can lead to misinterpretation of trends or incorrect model specifications that can negatively affect financial and policy decision-making process.

\subsection{Positioning of this study}
Existing studies support ensemble mortality forecasting and LLM application, but there is limited work combining both in an architecture. This project aims to fill in the gap by making the LLM a constrained orchestrator, while all forecasts and evaluation remain generated by validated actuarial models. The proposal's three strands are reflected end-to-end: a baseline evaluation under fixed baseline JSON and the same R pipeline, forecasting on the same executable path via run mode and effective-config fields; and a front-end (Streamlit over FastAPI) for submission, monitoring, and artefact access.

\section{Methodology}
This section explains our research design to investigate whether a large language model (LLM) can make CoMoMo-based mortality forecasting more usable without impacting statistical rigor. The design is mixed-methods in architecture but quantitative in evaluation: the LLM is studied as an orchestration and configuration component; all mortality estimates are produced by a fixed, auditable R pipeline using CoMoMo package, which is aligned with stacked-regression ensemble practice (Kessy et al., 2022). Other implementation details, such as API and user interface support the design but are subordinate to data choice, model set, validation protocol, metrics, and explicit assumptions.

\subsection{Research design}
Methodology has three main phases. In the first phase, we establish LLM to get natural language query as an input and then Kessy-aligned back testing and ensemble evaluation using the CoMoMo package in R to reproduce the male UK experiments. This shows that the code quality is guaranteed as well as LLM generated prompt can be mapped correctly to CoMoMo package which is illustrated by generating statistical outputs and graph visualisation. Following that, in the next stage, we extend that same pipeline to multi-step-ahead forecasting via explicit forecast horizons and run modes to produce predictions for unseen data. In the final phase, we wrap the workflow in a front-end interface so runs can be launched and monitored without handediting scripts. Python (LLM extraction, JSON validation, API, UI client) only prepares and transports configuration; all estimation, combination weights, MSE computation, and rate forecasts are produced in R.

\headingbf{Phase 1: Baseline replication and accuracy with CoMoMo}
A deterministic pipeline reproduces the model-combination and rolling-evaluation logic motivated by Kessy et al. (2022). These results are comparable across runs and can be checked against published ensemble methodology. Thus, this stage ensures computational accuracy and reproducibility before any natural-language layer is trusted.

\headingbf{Phase 2: Forecasting}
In this phase, we extend into forecasting. By fitting into the real data, it can produce the desired future year's prediction. The same R entry script extends the evaluation configuration to multi-step-ahead future rates, driven by run mode and forecast fields in the effective configuration. This avoids a separate ``forecast-only'' codebase that could drift from the evaluation implementation.

\headingbf{Phase 3: Interface and LLM-assisted configuration}
A web demonstration and optional local LLM extraction test whether non-programmers can launch valid runs; reliability is measured by valid merged configuration, successful completion, and complete artefacts, not by treating the LLM as a mortality model.

\subsection{Evaluation Strategy}
The evaluation of this study is structured around three key dimensions: correctness, reproducibility, and usability. First, correctness is evaluated by comparing the outputs of LLM-generated configurations with baseline CoMoMo results using rolling-origin mean squared error (MSE) and model ranking metrics. This ensures that the integration of the LLM does not compromise the statistical accuracy of the forecasting pipeline. Second, reproducibility is evaluated by enforcing fixed configurations and a deterministic modelling pipeline, allowing identical results to be produced across repeated runs under the same conditions. Finally, usability is examined through the system's ability to translate natural-language inputs into valid forecasting workflows and generate interpretable outputs. Together, these criteria provide a comprehensive framework for assessing both the technical reliability and practical effectiveness of the proposed LLM-integrated interface.

\subsection{Data source}
Data are drawn from the Human Mortality Database (HMD) as it is the reliable reference for international age-specific death rates. It is widely adopted in GAPC and ensemble mortality studies, and necessary for direct comparison with CoMoMo-style workflows in the literature.

For representative replication runs, the baseline effective configuration specifies England \& Wales aggregate (GBRTENW), male, with ages 50--89. That band is appropriate for this project because it targets adult mortality where the six fitted stochastic models (LC, RH, APC, CBD, M7, Plat) are routinely applied in actuarial and demographic work. More specifically, fit period 1960-1990 for metadata and weight-learning windows used in rolling evaluation; rolling origins 1990-2015 for out-of-sample testing; mse\_horizon\_max = 15 as the maximum forecast lead for weight estimation and MSE reporting; forecast\_max\_year = 2016 for the historical span referenced in the template.

\section{Results and Discussion}

\subsection{Baseline and Model Performance}
The baseline replication was reproduced for England and Wales (male) shows the reliability of the pipeline. This has leveraged the CoMoMo package compared to the intensive initial pipeline and set foundation for forecasting and interface integration. The figures below present the CVMSE and mean squared error (MSE) by forecast horizon for individual models.

\begin{figure}[t]
\centering
\includegraphics[width=0.48\linewidth]{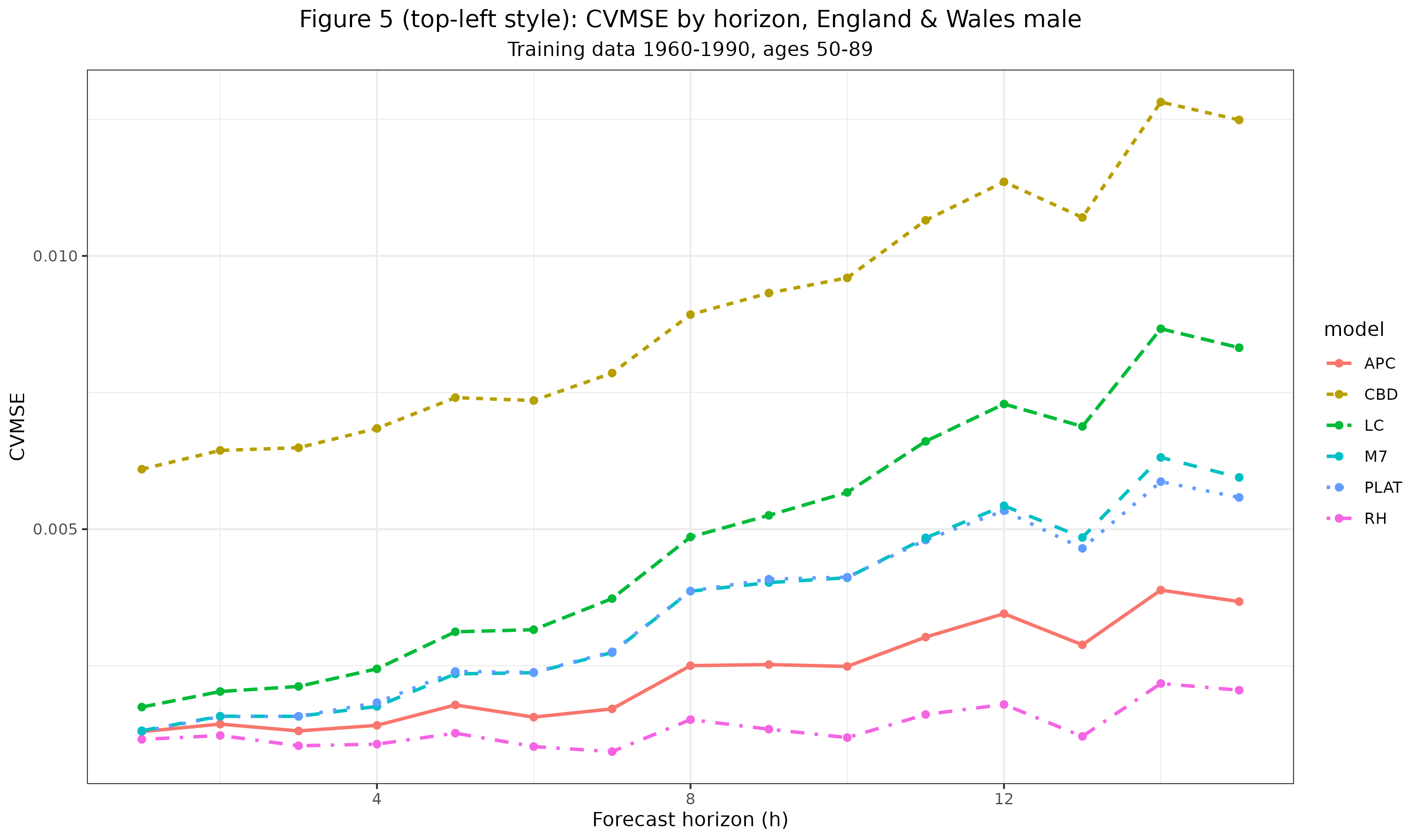}\hfill
\includegraphics[width=0.48\linewidth]{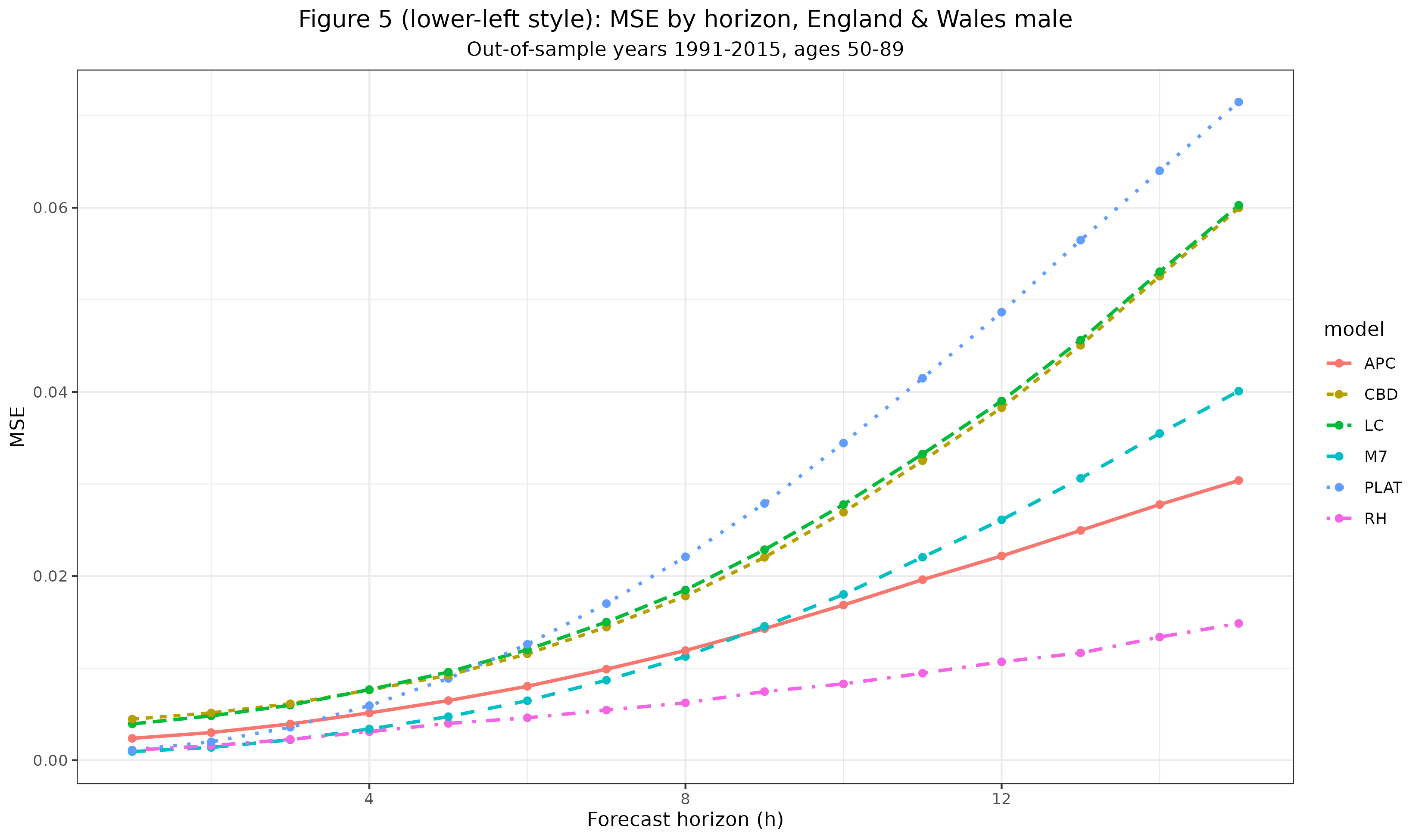}
\caption{Cross-validated mean squared error (CVMSE) and mean squared error (MSE) by forecast horizon for individual mortality models under rolling-origin evaluation, England and Wales (male population).}
\end{figure}

\begin{figure}[t]
\centering
\includegraphics[width=\linewidth]{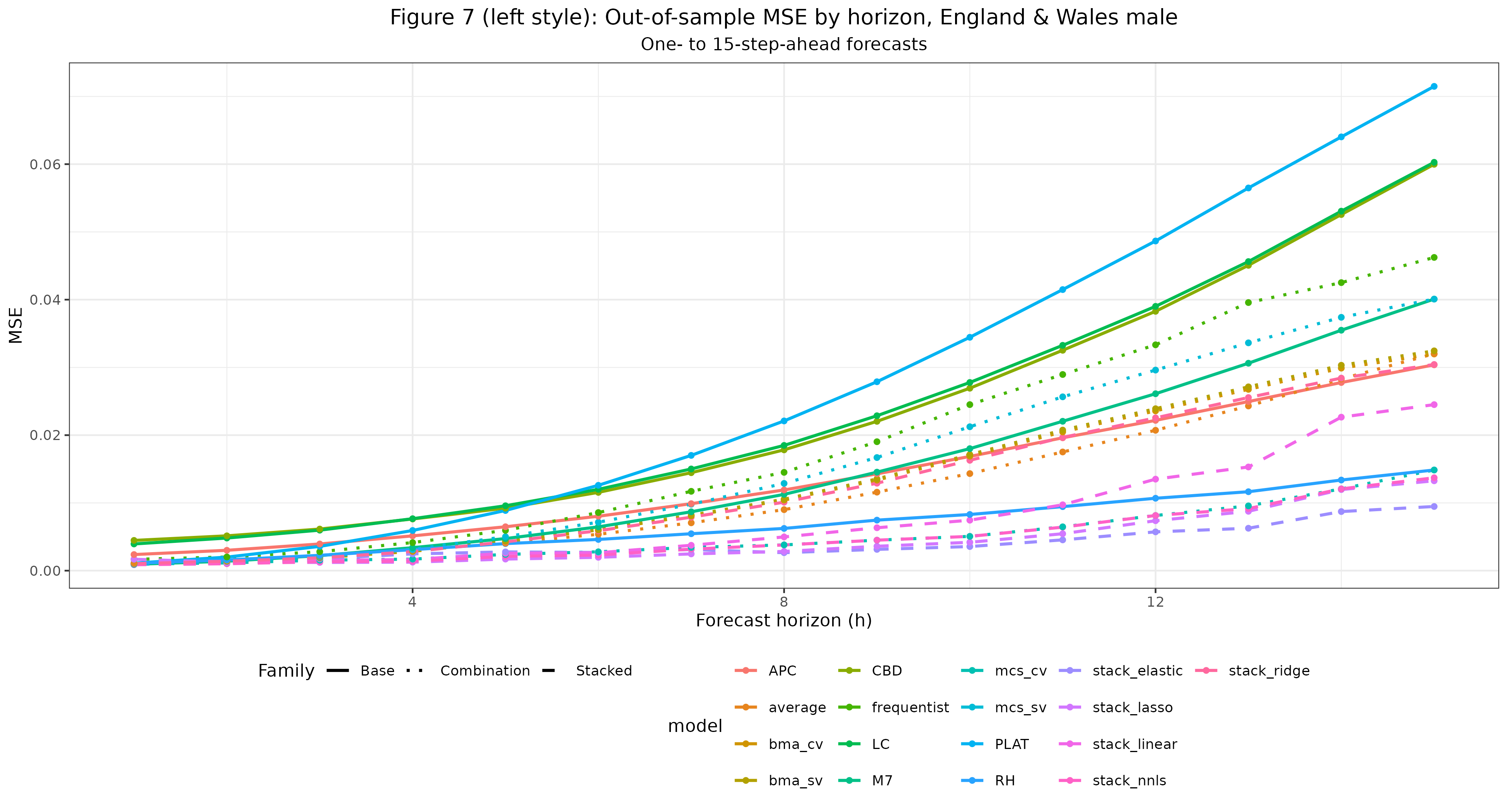}
\caption{Out-of-sample mean squared error (MSE) by forecast horizon for individual and ensemble mortality models under rolling-origin evaluation, England and Wales (male population).}
\end{figure}

As illustrated in the figures above, stacked regression approaches and other combination methods achieve lower MSE across nearly all forecast horizons in out-of-sample distribution.

\begin{figure}[t]
\centering
\includegraphics[width=\linewidth]{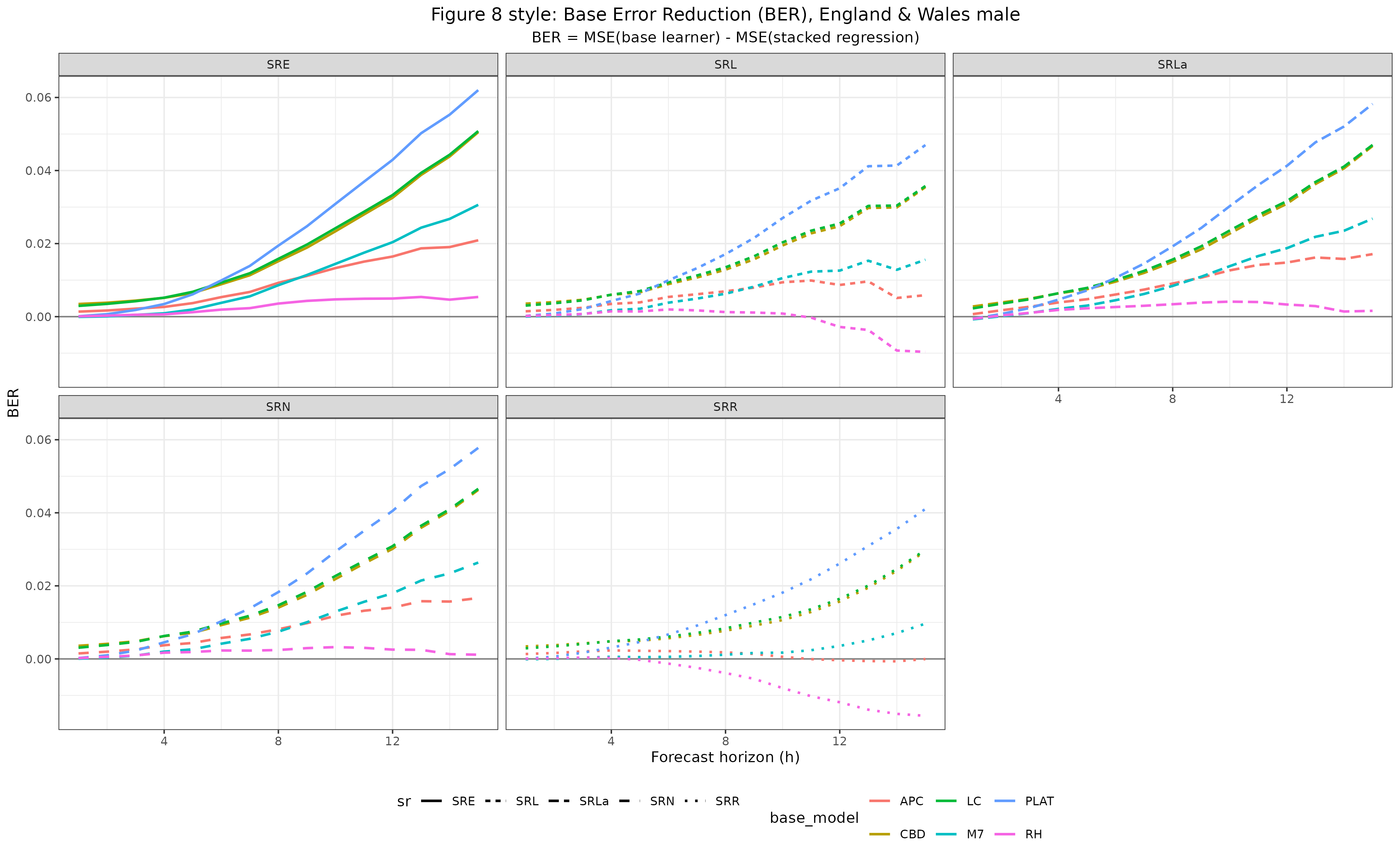}
\caption{Base error reduction (BER) of ensemble methods relative to individual mortality models across forecast horizons, England and Wales (male population).}
\end{figure}

\subsection{Forecasting Outputs and Practical Implications}
The extended forecasting process generates subsequent multi-step mortality rate forecasts under consistent evaluation conditions. The results show that the aggregated forecasts maintain continuity across all time periods, ensuring that comparative analyses between models remain understandable. From a practical perspective, this consistency is crucial for actuarial applications, where decision-makers rely on stable and comparable forecasts for pricing, reserves, and policy design. The ability to generate complete and coherent forecast trajectories reduces the risk of misinterpretation and supports more informed decision-making.

\begin{figure}[t]
\centering
\includegraphics[width=\linewidth]{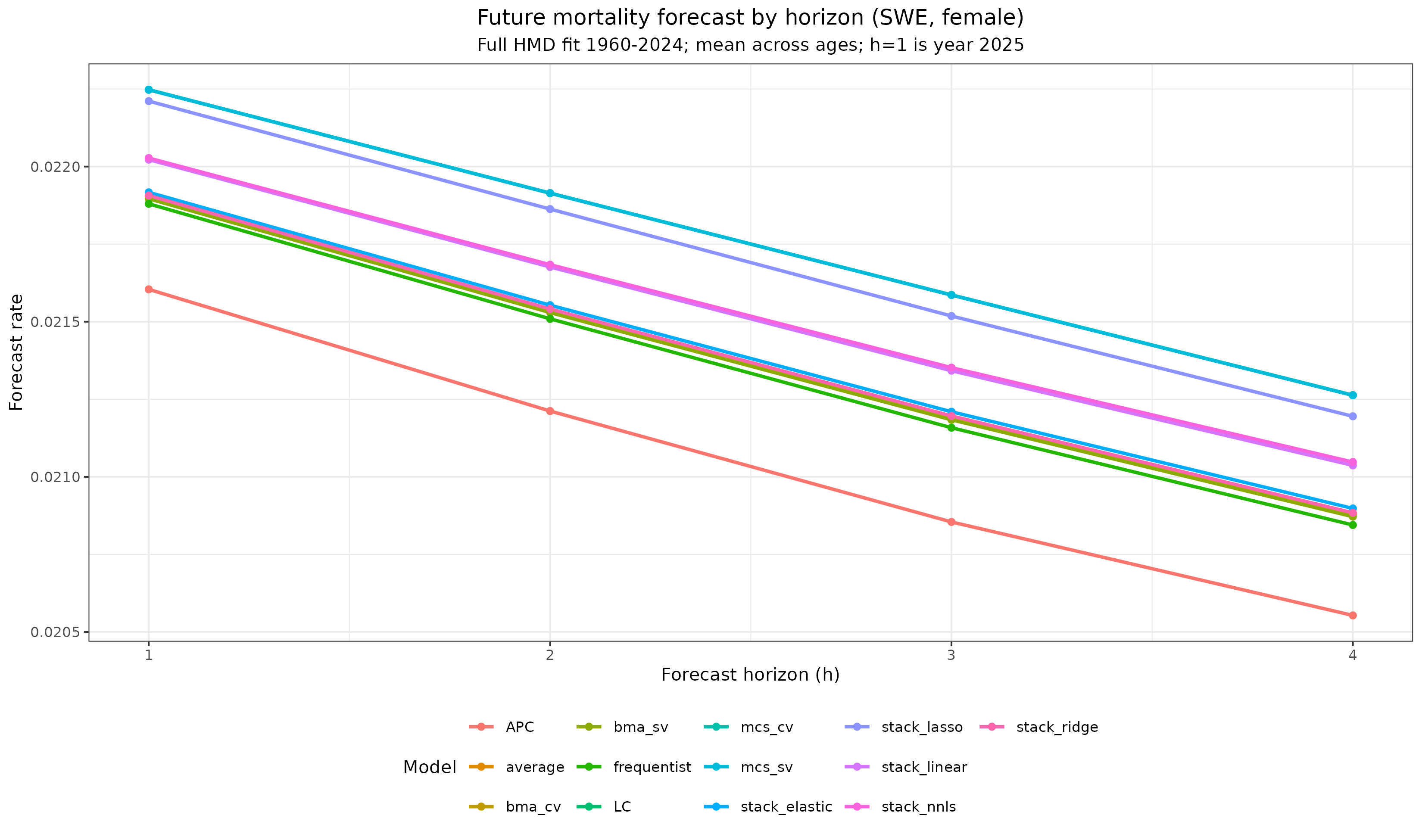}
\caption{Multi-step-ahead mortality forecasts by individual and ensemble models for Swedish female population under consistent evaluation settings.}
\end{figure}

\subsection{Interface and System Demonstration}
The prototype interface enables users to initiate and monitor forecasting runs without interacting directly with R code. The system successfully separates user interaction from statistical estimation, maintaining a clear boundary between input interpretation and model execution. This allows them to customise and extract mortality rates of different countries, genders, single models and meta-learners without having prior knowledge of the coding. From a usability perspective, this significantly reduces the technical barrier for non-expert users. At the same time, the architecture preserves transparency through artefact generation, execution logs, and reproducible configuration files. This balance between accessibility and auditability is a key contribution of the project.

The workflow has two different layers which are users' intent and statistical estimation. When users submit a natural-language request (or when LLM extraction is disabled a structured JSON string) together with run settings such as baseline configuration and run mode. The local large language model converts the prompt into a userrequest JSON object under a system prompt, and a validation step will then merge that object with a committed baseline effective configuration to build an effective configuration file that fully meet the requirements of the mortality study. In other words, they must have population, windows, model set, meta-learners, horizons. That file is passed to a deterministic R pipeline based on CoMoMo, which downloads Human Mortality Database data as required, fits the chosen stochastic models, runs rolling-origin evaluation (MSE by horizon and average rankings), and, when configured, generates future mortality rate forecasts. All numerical outputs are written to a timestamped run folder. A small HTTP API orchestrates these steps asynchronously, which enforces one active run at a time, streaming execution logs, and surfacing completion status and error, at the same time, a web front end lets users launch runs and inspect results. The LLM therefore acts only as a configuration translator when reproducibility and actuarial credibility rest on the validated JSON and the fixed R implementation.

The demonstration front end is a Streamlit application that acts as a thin client to the orchestration API. On load, it checks API health and presents sidebar controls for baseline configuration together with a text area for the user's request. Pressing Run Pipeline sends an authenticated POST request with the assembled payload; on success the UI stores the returned run identifier, clears any cached ZIP from a previous run, records a short confirmation message, and reruns the page so the display switches cleanly to the new job. While status remains queued or running, the app polls the API on a short interval and shows status, metadata, and execution logs. When the run completes, it fetches previews of MSE and forecast tables and offers Prepare ZIP / Download ZIP for the full output directory via a separate download endpoint with a longer timeout. Environment variables configure the API base URL and API key, so the same UI can target a local or shared backend without code changes.

\begin{figure}[t]
\centering
\includegraphics[width=\linewidth]{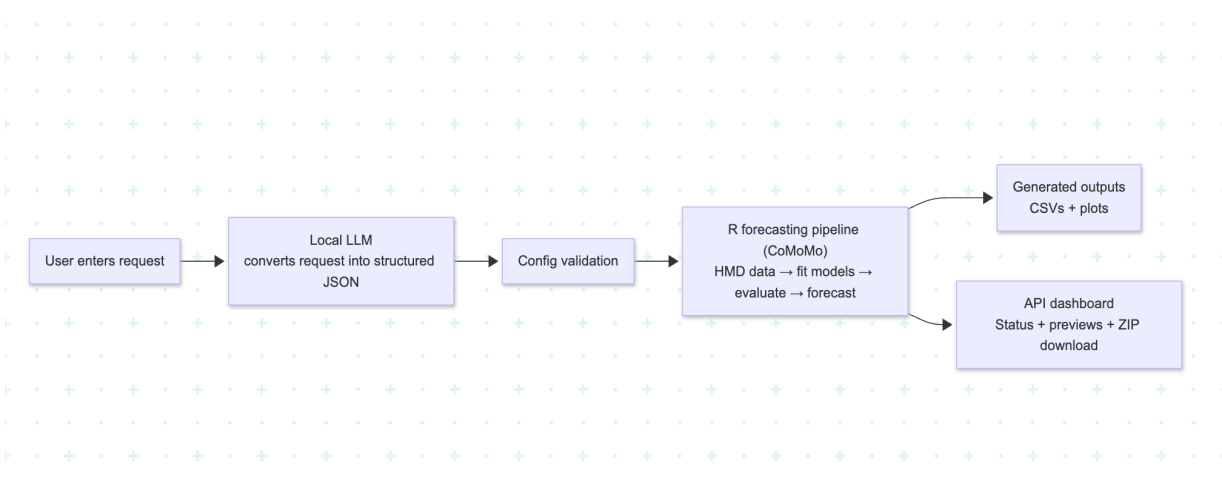}
\caption{Architecture of the LLM-integrated mortality forecasting pipeline, illustrating the flow from natural-language input to configuration validation and deterministic R-based model execution.}
\end{figure}

\begin{figure}[t]
\centering
\includegraphics[width=0.48\linewidth]{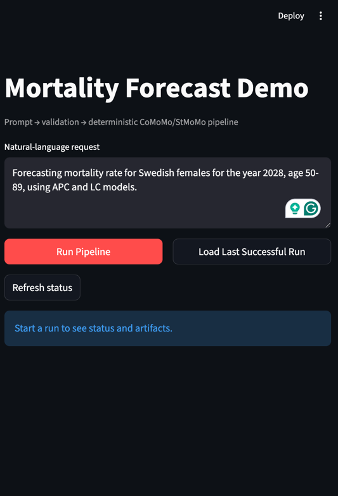}\hfill
\includegraphics[width=0.48\linewidth]{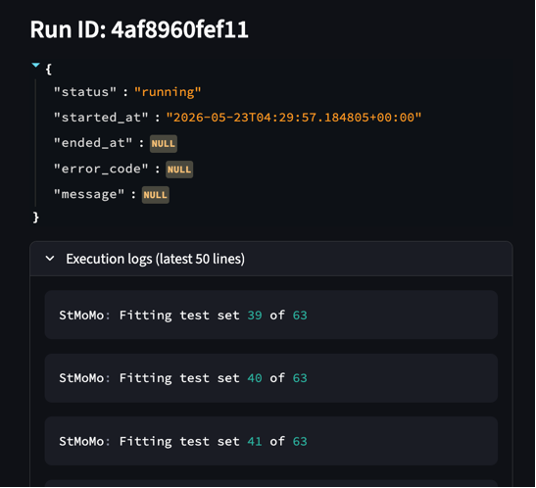}
\caption{Streamlit-based user interface for submitting natural-language mortality forecasting requests and initiating pipeline execution.}
\end{figure}

\begin{figure}[t]
\centering
\includegraphics[width=0.32\linewidth]{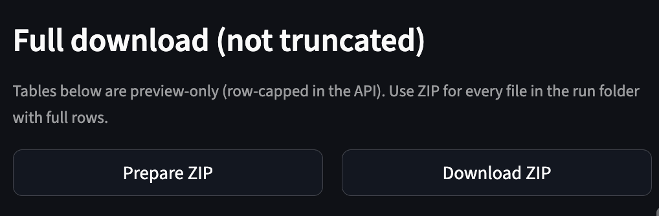}\hfill
\includegraphics[width=0.32\linewidth]{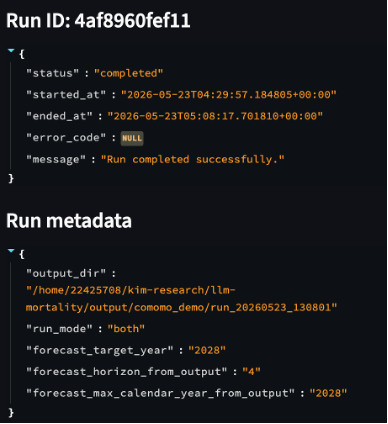}\hfill
\includegraphics[width=0.32\linewidth]{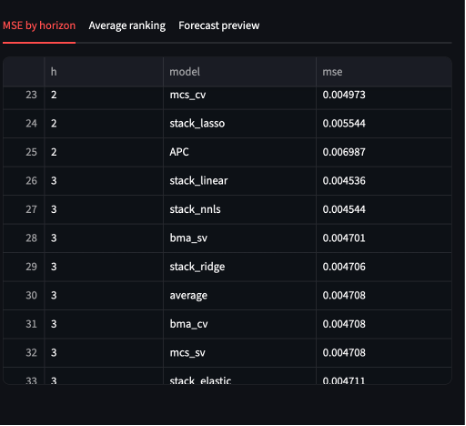}
\caption{Execution status and real-time logging of forecasting run within the user interface.}
\end{figure}

\subsection{Key Findings}
Overall, the results provide strong evidence that emphasising the ensemble methods outperform individual mortality models, particularly at longer horizons and deterministic pipelines ensure reproducibility and consistent evaluation. Moreover, LLMs can improve accessibility when constrained to orchestration roles and a hybrid architecture combining statistical model with natural-language interaction is both feasible and effective.

These findings directly address the research objectives of improving usability while maintaining correctness, reproducibility, and transparency in mortality forecasting workflows.

\section{Limitations}
This study still meets some challenges. First, the mortality forecasting pipeline requires computational resourced due to heavy model training process, ensemble generation and rolling-origin evaluation. Those processes are costly in mortality forecasting research and may reduce efficiency and scalability for practical employment. Another limitation is the use of local large language model (Llama). Though the local setup is helpful in security and privacy, smaller local models have limited interpretative capability compared to API-based models (e.g. OpenAI). Their weaker contextual understanding and less robust adherence to complex structured outputs can increase validation failures. Finally, the evaluation scope remains limited in terms of real-world generalisability. The experiments were conducted within a restricted setting, which expanding into different demographic groups, and forecasting contexts is still required.

\section{Future Work}
As the goal of the study is to narrow the gap between the mortality forecast modelling and non-technical user approach, future work would focus on improving the efficiency and scalability of the pipeline. We are planning to optimise the training progress and store it in cache to get ready for users' extraction. Regarding large language models component, it is recommended to investigate hybrid approaches that combine local models with stronger API-based models for more complex interpretation tasks. For instance, retrieval-augmented prompting, fine-tuning on validated prompts, and improved structured-output validation may further enhance robustness and reduce configuration failures. In addition, multi-country and multi-sex studies with stratified reporting would provide better understanding regarding the effectiveness and reliability of ensemble forecasting approaches.

\section{Conclusion}
This project aimed to design a reliable LLM-integrated interface to improve the usability of mortality forecasting while preserving statistical rigour, reproducibility, and transparency. The results demonstrate that this objective has been successfully achieved as a proof-of-concept that it is promising to apply natural-language interaction to enhance the approach of deterministic actuarial modelling. However, future work such as optimising the training process of the model or enhancing interpretability of LLM should be taken into consideration to be scalable and ready for deployment. Overall, this study contributes a practical and scalable design pattern for integrating LLMs into high-stakes analytical workflows, with direct relevance to actuarial practice and data-driven decision-making.

\nocite{*}
\bibliographystyle{aaai}
\bibliography{references}

@techreport{bommasani2021,
  author      = {Bommasani, Rishi and Hudson, Drew A. and Adeli, Ehsan and others},
  title       = {On the Opportunities and Risks of Foundation Models},
  institution = {Stanford University Center for Research on Foundation Models},
  year        = {2021}
}

@inproceedings{brown2020,
  author    = {Brown, Tom B. and Mann, Benjamin and Ryder, Nick and others},
  title     = {Language Models are Few-Shot Learners},
  booktitle = {Advances in Neural Information Processing Systems},
  volume    = {33},
  year      = {2020}
}

@article{bubeck2023,
  author  = {Bubeck, S{\'e}bastien and Chandrasekaran, Varun and Eldan, Ronen and others},
  title   = {Sparks of Artificial General Intelligence: Early Experiments with {GPT-4}},
  journal = {arXiv preprint arXiv:2303.12712},
  year    = {2023}
}

@article{cairns2006,
  author  = {Cairns, Andrew J. G. and Blake, David and Dowd, Kevin},
  title   = {A Two-Factor Model for Stochastic Mortality},
  journal = {Journal of Risk and Insurance},
  volume  = {73},
  number  = {4},
  pages   = {687--718},
  year    = {2006}
}

@misc{dattani2023,
  author       = {Dattani, Saloni and Rod{\'e}s-Guirao, Lucas and Ritchie, Hannah and Ortiz-Ospina, Esteban and Roser, Max},
  title        = {Life Expectancy},
  howpublished = {Our World in Data},
  year         = {2023},
  url          = {https://ourworldindata.org/life-expectancy},
  note         = {Accessed 2026-05-30}
}

@book{hyndman2018,
  author    = {Hyndman, Rob J. and Athanasopoulos, George},
  title     = {Forecasting: Principles and Practice},
  edition   = {2},
  publisher = {OTexts},
  address   = {Melbourne, Australia},
  year      = {2018},
  url       = {https://otexts.com/fpp2/}
}

@inproceedings{jacovi2020,
  author    = {Jacovi, Alon and Goldberg, Yoav},
  title     = {Towards Faithfully Interpretable {NLP} Systems: How Should We Define and Evaluate Faithfulness?},
  booktitle = {Proceedings of the 58th Annual Meeting of the Association for Computational Linguistics},
  pages     = {4198--4205},
  year      = {2020}
}

@article{kessy2022,
  author  = {Kessy, Samuel R. and Sherris, Michael and Villegas, Andr{\'e}s M. and Ziveyi, Jonathan},
  title   = {Mortality Forecasting Using Stacked Regression Ensembles},
  journal = {Scandinavian Actuarial Journal},
  volume  = {2022},
  number  = {7},
  pages   = {591--626},
  year    = {2022}
}

@article{lee1992,
  author  = {Lee, Ronald D. and Carter, Lawrence R.},
  title   = {Modeling and Forecasting {U.S.} Mortality},
  journal = {Journal of the American Statistical Association},
  volume  = {87},
  number  = {419},
  pages   = {659--671},
  year    = {1992}
}

@techreport{openai2023,
  author      = {{OpenAI}},
  title       = {{GPT-4} Technical Report},
  institution = {OpenAI},
  year        = {2023},
  note        = {arXiv:2303.08774}
}

@article{plat2009,
  author  = {Plat, Richard},
  title   = {On Stochastic Mortality Modeling},
  journal = {Insurance: Mathematics and Economics},
  volume  = {45},
  number  = {3},
  pages   = {393--404},
  year    = {2009}
}

@article{renshaw2006,
  author  = {Renshaw, Arthur E. and Haberman, Steven},
  title   = {A Cohort-Based Extension to the {Lee--Carter} Model},
  journal = {Insurance: Mathematics and Economics},
  volume  = {38},
  number  = {3},
  pages   = {556--570},
  year    = {2006}
}

@article{shang2018,
  author  = {Shang, Han Lin and Haberman, Steven},
  title   = {Model Selection and Averaging in Mortality Forecasting},
  journal = {Annals of Actuarial Science},
  volume  = {12},
  number  = {1},
  pages   = {1--27},
  year    = {2018}
}

@article{villegas2018,
  author  = {Villegas, Andr{\'e}s M. and Kaishev, Vladimir K. and Millossovich, Pietro},
  title   = {{StMoMo}: Mortality Stochastic Modelling in {R}},
  journal = {Journal of Statistical Software},
  volume  = {84},
  number  = {3},
  pages   = {1--38},
  year    = {2018},
  doi     = {10.18637/jss.v084.i03}
}

@inproceedings{wei2022,
  author    = {Wei, Jason and Wang, Xuezhi and Schuurmans, Dale and others},
  title     = {Chain-of-Thought Prompting Elicits Reasoning in Large Language Models},
  booktitle = {Advances in Neural Information Processing Systems},
  volume    = {35},
  year      = {2022}
}

@article{wolpert1992,
  author  = {Wolpert, David H.},
  title   = {Stacked Generalization},
  journal = {Neural Networks},
  volume  = {5},
  number  = {2},
  pages   = {241--259},
  year    = {1992}
}

@article{yao2018,
  author  = {Yao, Yuling and Vehtari, Aki and Simpson, Daniel and Gelman, Andrew},
  title   = {Using Stacking to Average Predictive Distributions},
  journal = {Bayesian Analysis},
  volume  = {13},
  number  = {3},
  pages   = {917--1007},
  year    = {2018}
}

\end{document}